\documentclass[11pt,a4paper]{article}
\usepackage{tcolorbox}
\usepackage{emnlp2021}
\usepackage{lipsum} 
\usepackage{times}
\usepackage{comment}
\usepackage{latexsym}

\usepackage{microtype}

\usepackage{subfiles}
\usepackage{graphicx}
\usepackage{caption}
\usepackage{subcaption}
\usepackage{rotating}
\usepackage{booktabs}
\usepackage{cleveref}
\usepackage{float} 
\usepackage{multirow}
\usepackage{soul}
\usepackage{color}
\newcommand\BibTeX{B\textsc{ib}\TeX}

\definecolor{bluehl}{RGB}{219,233,243}
\definecolor{orangehl}{RGB}{255,127,14}

\DeclareRobustCommand{\hlblue}[1]{{\sethlcolor{bluehl}\hl{#1}}}
\title{Hierarchical Learning for Generation with Long Source Sequences}

\author{Tobias Rohde$^{ \dagger 	\ddagger }$ Xiaoxia Wu$^{\dagger  }$\thanks{ { } Work performed while Xiaoxia Wu interned at Birch AI.  She is currently a postdoc fellow at The University of Chicago and Toyota Technological Institute at Chicago.} , Yinhan Liu$^{\dagger }$  \\
\\
  ${\dagger }$ Birch AI, Seattle, WA \vspace{0.2cm}\\
  ${\ddagger}$ Paul G. Allen School of Computer Science \& Engineering,\\
University of Washington, Seattle, WA\\
  \texttt{\{tobiasr, xwu, yinhan\}} @birch.ai\\
  }

\date{May 17th, 2021}
\usepackage{placeins}
\usepackage{url}
\begin{document}
\maketitle
\begin{abstract}
 One of the challenges for current sequence to sequence (seq2seq) models is processing long sequences, such as those  in summarization and document level machine translation tasks. These tasks require the model to reason at the token level as well as the sentence and paragraph level.
 We design and study a new Hierarchical Attention Transformer-based architecture (HAT) that outperforms standard Transformers on several sequence to sequence tasks. Furthermore, our model achieves state-of-the-art ROUGE scores
 on four summarization tasks, including PubMed, arXiv, CNN/DM, SAMSum, and AMI. Our model outperforms document-level machine translation baseline on the WMT20 English to German translation task.
 We investigate what the hierarchical layers learn by visualizing the hierarchical encoder-decoder attention. Finally, we study hierarchical learning on encoder-only pre-training and analyze its performance on classification tasks.
\end{abstract}

\section{Introduction}
\label{sec:intro}
Sequence to sequence (seq2seq) models have been successfully used for a variety of natural language processing (NLP) tasks, including text summarization,
machine translation and question answering. Often sequence to sequence models consist of an encoder that processes some source sequence and a decoder that
generates the target sequence. 
Originally, \citet{10.5555/2969033.2969173} used recurrent neural networks as encoder and decoder for machine translation on the WMT-14 dataset.
\citet{bahdanau2016neural} introduced the attention mechanism, where the decoder computes a distribution over the hidden states of the encoder and uses it
to weigh the hidden states of the input tokens differently at each decoding step.
\citet{vaswani2017attention} then introduced a new architecture for sequence to sequence modeling -- the Transformer, which is based on the attention mechanism but not recurrent allowing for more efficient training.

While successful, both recurrent neural networks and Transformer-based models have limits on the input sequence length. When the input is long, the learning degrades particularly for tasks which require a comprehensive understanding of the entire paragraph or document. One of the main learning challenges for seq2seq models is that the decoder needs to attend to token level representations from the encoder to predict the next token, while at the same time it must learn from a large context. 

A commonly used method for attempting to solve the long-sequence problem is hierarchical attention~\cite{yang2016hierarchical}. This method was studied primarily on long sequence classification tasks, where the model learns a document representation which is used as the input to a classifier. Since then, many successful
papers proposed methods using hierarchical attention (see Section \ref{sec:background} for full details).
While hierarchical attention has been successfully applied to classification tasks, its potential for being applied to large document sequence to sequence tasks remains an interesting and open question.

\begin{figure*}[h!]
\vspace{-0.4cm}
\centering
\includegraphics[width=\linewidth]{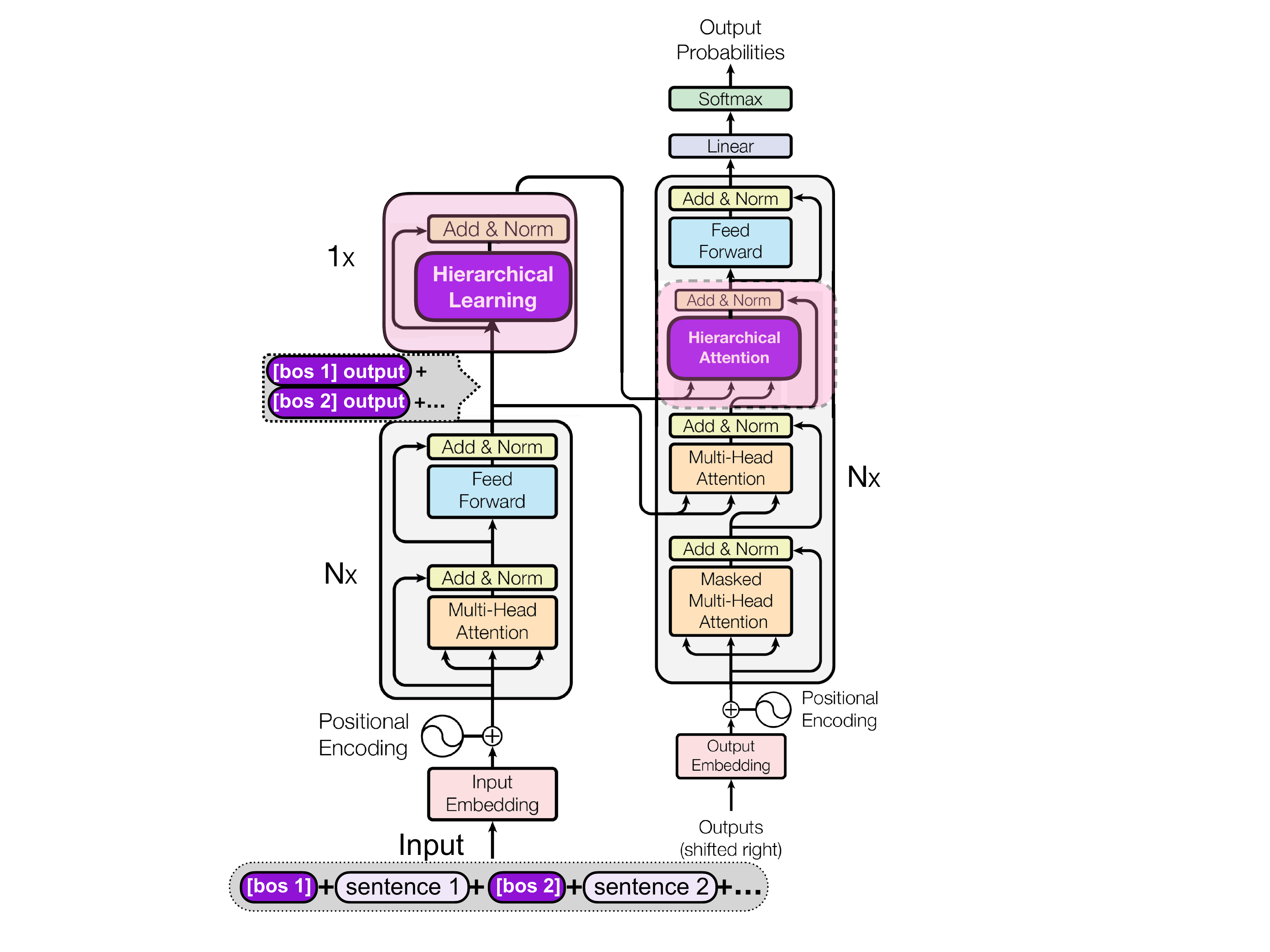}
 \vspace{-1.0cm}
          \caption{ \textbf{The architecture of HAT: Hierarchical Attention Transformer.} The blocks shaded in pink are the additional hierarchical layers added to the existing Transformer \cite{vaswani2017attention} architecture. The extra BOS tokens for hierarchical learning are highlighted in purple. The figure is based on Figure 1 in \citet{vaswani2017attention}.  }
          \label{fig:network}
 \vspace{-0.3cm}         
\end{figure*}

In this paper, we present a hierarchical attention model based on the standard Transformer~\cite{vaswani2017attention} that produces
sentence level representations, and combine them with token level representations to improve performance on long document sequence to sequence tasks. Our main contributions include
\begin{itemize}
    \item[1)] We design a hierarchical seq2seq attention network architecture named HAT (\textbf{H}ierachical \textbf{A}ttention \textbf{T}ransformer) (See Figure~\ref{fig:network}). We conduct extensive experiments on various generation tasks for our proposed model in Sectios~\ref{sec:experiment} and~\ref{sec:result} and achieve new state-of-the-art results on several datasets.
    \item[2)] In Sections~\ref{sec:analysis} and~\ref{sec:ablation}, we study the generated output of our architecture to further understand the benefits of hierarchical attention and compare it with those generated by plain seq2seq Transformer models. Furthermore, we analyze how
    the decoder makes use of the sentence level representations of the encoder by visualizing the hierarchical encoder-decoder attention.
    \item[3)] Finally, we apply hierarchical attention to an encoder-only architecture and pre-train it on a Books and Wiki corpus similar to the one used in RoBERTa~\cite{liu2019roberta}. We fine-tune our pre-trained model on several downstream tasks and analyze the performance in Section~\ref{sec:encoderonly}.
\end{itemize}

\section{Background}
\label{sec:background}

\textbf{Transformer models.} 
The attention-based Transformer architecture~\cite{vaswani2017attention} currently dominates the state-of-the-art performance in many NLP tasks.
This is largely due to the success of language model pre-training prior to fine-tuning the Transformer on the actual downstream task of interest.
Pre-training has been applied in different ways to different variations of the Transformer architecture, including
encoder-only pre-training (BERT~\cite{devlin2019bert}, RoBERTa~\cite{liu2019roberta}, XLNet~\cite{NEURIPS2019_dc6a7e65}), decoder-only pre-training (GPT~\cite{radford2019language}), encoder-decoder pre-training (T5~\cite{2020t5}, BART~\cite{lewis2020bart}) and multilingual pre-training (MBART~\cite{liu2020multilingual}, Roberta-XLM~\cite{conneau-etal-2020-unsupervised}). Both classification and generation downstream task performance is improved significantly by initializing parameters from pre-trained models. 
To use a pre-trained model, the downstream task model architecture needs to be the same or similar to the one used in pre-training.
Transformers have become the most popular architectures in NLP. However, one disadvantage of Transformers is that when the input sequence is long, the performance of the attention mechanism can become worse.


 \noindent\textbf{Long document modeling.}  Understanding and modeling large documents has been a longstanding challenge that has become increasingly demanded in practice~\cite{Nakao00,MihalceaC07,IyerKCZ16,ZhangSWGHC17,CelikyilmazBHC18,paulus2018a,TasnimCGOV19,ZhaoPFLLY19,gunel2019mind,ye2019bp,Beltagy2020Longformer,ZhangDLCZC20}. Many methods have been proposed to improve performance on abstractive summarization tasks, which often have long source sequences. Recent work includes graph-based attention neural networks~\cite{tan2017abstractive}, discourse-aware attention models~\citet{cohan-etal-2018-discourse} and decoder decomposition with policy learning~\citet{KryscinskiPXS18}. In addition, ~\citet{XuGCL20} suggest a discourse-aware neural summarization model with structural discourse graphs to capture the long-range dependencies among discourse units. ~\citet{zhu2020hierarchical} design a hierarchical network for meeting summarization that hierarchically processes speaker turns and incorporates speaker roles.

\noindent\textbf{Hierarchical learning.} Hierarchical learning has been suggested by many researchers and it has been empirically shown to be effective in numerous diverse tasks outside of natural language processing, including policy learning ~\cite{ryu2020multi}, visual relationship detection~\cite{mi2020hierarchical}, part-aware face detection~\cite{WuKSC19}, visually-aware food recommendation~\cite{GaoFHHGFMC20}, urban anomaly prediction~\cite{HuangZDB20}, online banking fraud detection~\cite{AchituveKG19} and discourse parsing~\cite{LiLC16}. Within the NLP domain,
hierarchical learning has been successfully applied to text classification \cite{yang2016hierarchical}, machine translation \cite{miculicich-etal-2018-document}, meeting summarization \cite{zhu2020hierarchical} and 
information extraction \cite{gao2018hierarchical,SongXWWZZ017,xing2018hierarchical,ZhaoJWC18,HanYLSL18}.
Particularly, \citet{yang2016hierarchical} propose a hierarchical structure with word and sentence-level attention mechanisms for document classification. \citet{xing2018hierarchical} design a hierarchical recurrent attention network to model the hierarchical structure of context, word importance, and utterance importance.  \citet{gao2018hierarchical} show that hierarchical attention networks perform significantly better than conventional models for information extraction from cancer pathology reports. 
\citet{miculicich-etal-2018-document} develop a hierarchical attention network that improves machine translation performance
by integrating information from previously translated sentences within a document.

In this work, we apply hierarchical learning to Transformer models for improving performance on 
generation tasks with long documents, including summarization and document-level machine translation.
\section{Model}
\label{sec:model}


We modify the standard sequence to sequence transformer architecture~\citep{vaswani2017attention} by adding hierarchical attention for improved processing of long documents (Figure \ref{fig:network}).  The number of parameters for our large hierarchical model on summary tasks is 471M compared to the plain transformer 408M.

We use 12 encoder and decoder layers, a hidden size of 1024, 4096 for the dimension of the fully-connected
feed-forward networks and 16 attention heads in both the encoder and the decoder.
Unlike the original Transformer, we use GELU activation instead of ReLU. 

\subsection{Data pre-processing}
During pre-processing, we insert BOS tokens at the start of every sentence in each source document as shown in Figure~\ref{fig:network}. We simply
determine sentence boundaries by punctuation or by using prior sentence segmentation present in the documents.
By using BOS tokens as hierarchical tokens, the hierarchical attention can benefit from the representations learned for the BOS token during pre-training.
\subsection{Encoder}
We use the same encoder as in Transformer. This produces an embedding for each input token.
After those, we add an additional encoder layer (pink block in Figure~\ref{fig:network}) which only attends to the embeddings of the BOS tokens that we inserted during data-preprocessing. We refer to this layer as the hierarchical encoder layer, which produces another level of contextual representations for each of the BOS tokens, which can be interpreted as sentence level representations. We find that a single hierarchical layer works the best,
although multiple hierarchical layers may be used.
\subsection{Decoder}
As in \citet{vaswani2017attention}, each layer first performs self attention over the previously generated tokens and then attends over the outputs of the final token level encoder layer, similarly to the vanilla Transformer. We add an attention module that attends over the BOS token embeddings from the hierarchical encoder layer.
\section{Experiments}
\label{sec:experiment}

Our architecture is specifically designed to better handle long sequences, thus we evaluate it
on summarization and document level translation tasks, which tend to have long source sequences as the input.
We also run experiments with non-generation tasks with an encoder-only hierarchical attention architecture (Section \ref{sec:encoderonly}).

\subsection{Summarization Tasks}

We characterize all the summarization tasks into several categories, test our architectures with different weight initializations and compare them with their non-hierarchical counterparts using the same weight initializations. 

\paragraph{Long sequence datasets}
The PubMed and arXiv datasets~\cite{cohan-etal-2018-discourse} contain scientific articles
from PubMed and arXiv respectively, and use the abstract of the articles as the target summary.
The sequences in each of these datasets are long and need to be truncated to be processed by most transformers. Statistics on the PubMed and arXiv datasets are given in Table~\ref{table:datastats}.
\begin{table*}[h!]
\centering
\begin{tabular}{c c c c c c c c} 
\toprule
{}
&\multicolumn{3}{c}{\textbf{Instances}} & \multicolumn{2}{c}{\textbf{Input Length}} &
\multicolumn{2}{c}{\textbf{Output Length}} \\
\hline
Dataset & Train & Valid & Test & Words & Tokens & Words &  Tokens\\
\hline
PubMed &112K & 6.6K & 6.7K & 3016 & 4016 & 178  & 258\\ 
ArXiv & 203K & 6.4K & 6.4K & 6055 & 8613 & 280 & 362\\ 
CNN-DM & 287K & 13K & 11K & 781 & 906 & 56 & 63\\ 
XSum & 204K & 11K & 11K & 431 & 488 & 23 & 27\\ 
SAMSum & 14.7K & 818 & 819 & 94 & 147 & 20 & 26 \\
AMI & 100 & 17 & 20 & 3156 & 4081  & 280 & 321\\
ISCI & 43  & 10 & 6 & 6228 &7913  &466 & 576\\
\hline
\end{tabular}
\caption{Stats for each of the summarization datasets. Both average number of words and number of BPE tokens are presented. Words are counted before tokenization.}
\label{table:datastats}

\centering
\begin{tabular}{c c c c c c c c} 
\toprule
{}
&\multicolumn{3}{c}{\textbf{Documents}} & \multicolumn{2}{c}{\textbf{Source Length}} &
\multicolumn{2}{c}{\textbf{Target Length}} \\
\hline
Dataset & Train & Valid & Test & Words & Tokens & Words &  Tokens\\
\hline
WMT20 En-De & 363K & 122 & 130 & 354 & 446 & 331 & 460 \\
WMT20 En-Cs & 42K & 107 & 130 & 738 & 903 & 641 & 862 \\
TED17 Zh-En & 1906 & 97 & 12 & 3775 & 2680 & 2107 & 2590 \\
\hline
\end{tabular}
\caption{Statistics for the WMT20 En-De, WMT20 En-Cs and TED17 Zh-En translation datasets. Both average number of words and number of BPE tokens are presented.}
\label{table:mt-datastats}
\end{table*}

We add a BOS token at the beginning of each sentence in the source sequences during data preprocessing. We use 3072 as the maximum source length, 512 as the maximum target length. Longer sequences are truncated. We follow BART~\cite{lewis2020bart} and use GPT2~\cite{radford2019language} byte-pair encoding (BPE). We random initialize the hierarchical encoder layer, the hierarchical attention modules in the decoder layers and the additional positional embedding from 512 to 3072. We initialize all the remaining weights with pre-trained weights from BART. We also initialize a plain seq2seq model with BART pre-trained weights for direct comparison.

For training, we use a batch-size of 128. We set weight decay to 0.01 and use the Adam optimizer with $(\beta_1, \beta_2)=(0.9,0.999)$ and $\varepsilon=10^{-8}$ \cite{KingmaB14}. We train for 30000 steps and warm up the learning rate for 900 steps to $3 \times 10^{-5}$ and decay it linearly afterwards. We use a dropout rate of 0.1 for attention and all layers. We also use label smoothing with smoothing constant 0.1. We use mixed precision for training. We complete 30000 steps in approximately 115 hours (100 hours) for HAT (plain-transformer) on arXiv with 2 A100 GPUs.\footnote{Only training time excluding validation and model saving.} We train three models with different batch sizes (32, 64, and 128) and choose the best model based on minimizing label smoothing loss. We used the same training parameters for both plain and hierarchical models.

For generation, we use a beam width of 2, length penalty of 1 and minimum and maximum generation lengths of 72 and 966 respectively.


\paragraph{News datasets}
CNN/DailyMail~\cite{NIPS2015_afdec700} and XSum~\cite{narayan-etal-2018-dont} are commonly used news summarization datasets. Both datasets are sourced from news articles, which frequently include a short summary in the article. Statistics for CNN/DM and Xsum are in Table~\ref{table:datastats}.

We use 1024 as maximum source length for article and 256 as maximum target length for the summary. Longer sequences are truncated. We apply the same data-processing and model initialization for these two datasets as we did for the long sequence datasets. We train on CNN/DM for 20000 steps with a batch-size of 64 and a peak learning rate of $3 \times 10^{-5}$ with linear decay afterwards. During generation, we use beam size 1, no length penalty and minimum and maximum generation lengths of 40 and 120, respectively. We use the same training and generation parameters for both the hierarchical seq2seq model and the plain seq2seq model. We grid search on batch sizes (32 and 64) and number of total steps (20k and 30k) and choose the best model by label smoothing loss. 

\paragraph{Conversational datasets}
Since conversational summarization datasets are rare, we only consider the SAMSum~\cite{gliwa-etal-2019-samsum} corpus.
It consists of staged chat conversations between two people and corresponding abstractive summaries written by linguists. Statistics for SAMSum datset are presented in Table~\ref{table:datastats}.

During data-processing, we concatenate the role string with its utterance and add a BOS token at the beginning of each speaker turn. Then we concatenate all the turns together and use this as the source sequence. We add segmentation embeddings to our model, where we map the same role along with its utterance to the same segment embedding. The input to the encoder layers is the sum of the token embeddings, position embeddings and segment embeddings. We randomly initialize segment embedding parameters, and initialize the remaining parameters with the hierarchical seq2seq model trained on CNN-DM. For comparison, we also use train a plain seq2seq model initialized with weights from a plain seq2seq trained on CNN-DM.

\paragraph{Meeting datasets}
The AMI~\cite{f0d6070c879244468a1e4ef5c5a1b81f} and ISCI~\cite{Janin03theicsi} corpora consist of staged 
meetings which are transcribed and annotated with abstractive
summaries. The meeting transcripts are extremely long and the turn-based structure of the meetings makes 
these datasets particularly suitable for the hierarchical architecture, since speaker turns can be marked by hierarchical tokens. Statistics for AMI and ISCI datasets are illustrated in Table~\ref{table:datastats}. We followed~\cite{shang-etal-2018-unsupervised} for splitting the data.

Since the meeting transcripts are transcribed from a speech to text model, we first built a small lexicon that filters out meaningless words. Then we add BOS tokens to each turn and concatenate all the turns together, which we use as input to the model. Following the conversational dataset approach, we add segment embeddings for each role. We follow the same weight initialization procedure as with the conversational datasets.

\subsection{Document level Machine Translation.}
Historically, sentence level translation approaches have outperformed document level approaches for translating entire documents. This indicates that document level
approaches are not able to incorporate the larger context of the document during translation. We test our hierarchical model on a translation task and see significant improvements
over the non-hierarchical counterpart on English to German document translation.
\paragraph{Datasets}
We evaluate our architecture on the WMT20 English to German, English to Czech and TED17 Chinese to English translation tasks.
Dataset statistics are shown in Table~\ref{table:mt-datastats}. We only
process 512 tokens at once due to memory constraints. Thus we split documents into chunks of at most 512 tokens. We only split at segment boundaries
to maintain alignment between the source and target languages. We translate each chunk separately and afterwards concatenate the translated chunks
to compute the BLEU score for the entire document.
We use moses for preprocessing the English, German and Czech text and Jieba for tokenizing the Chinese text. We use the fastBPE implementation of \citet{sennrich-etal-2016-neural} for byte-pair encoding with 40000 bpe codes and joined dictionaries
for all tasks. For the WMT20 tasks, we use the WMT18 test data for validation and test the models on both the WMT19 and WMT20 test data. For TED17, we use the test and validation data prior to 2016 as validation data and test the model on the
TED16 and TED17 test data.

\paragraph{Model and Optimization}
For English to German and English to Czech translation, we use the Transformer architecture \citep{vaswani2017attention} with 6 encoder/decoder layers, a hidden size of 1024, 16 attention heads and 4096 for the dimension
of the feedforward networks. The hierarchical model has 260M parameters and the plain model has 222M parameters.

For Chinese to English translation we use the same architecture, but with a hidden size of 512, 4 attention heads and 1024
for the dimension of the feedforward netoworks. For the hierarchical models, we use one hierarchical encoder layer.
The hierarchical model has 64M parameters and the plain model has 55M parameters.

For all training experiments we use the Adam optimizer with $(\beta_1, \beta_2) = (0.9, 0.98)$ and $\varepsilon = 10^{-6}$.
For training English to German translation, we use 1024 tokens per batch, 16 gradient accumulation steps and we train on 8 V100
GPUs.  We minimize the label smoothed cross entropy loss with smoothing constant 0.1 and use dropout of $0.1$ for regularization. We train with mixed precision for a total of 50000 steps and warm up the learning rate for 1250 steps to $10^{-4}$ and decay it linearly afterwards. Training takes about 25 hours
for the hierarchical model and 17 hours for the plain model.

For training English to Czech translation, we use $1024$ tokens per batch, 16 gradient accumulation steps and we train on
8 V100 GPUs. We use label smoothing with smoothing constant $0.2$, a dropout of $0.2$ and $0.1$ weight decay.
We train with mixed precision for a total of 15000 steps and warm up the learning rate to $5 \times 10^{-4}$ for 500 steps and decay it linearly to $0$ afterwards. Training takes about 8 hours for the hierarchical model and 5 hours for the plain model.

For training English to Chinese translation, we use $8192$ tokens per batch, 2 gradient accumulation steps and we train on one
V100 GPU. We use label smoothing with smoothing constant $0.2$, a dropout of $0.3$ and $0.1$ weight decay.
We train with mixed precision for a total of $8000$ steps and warm up the learning rate to $10^{-3}$ for $100$ steps and decay it linearly to $0$ afterwards.
Training takes about 1.5 hours for the hierarchical model and 1 hour for the plain model.

For all languages, we use a beam width of $4$ during generation and we generate until
we encounter an EOS token. We manually tuned these parameters by choosing the best validation BLUE score. However, note that the above parameters were not extensively tuned and we do not use a pre-train.
\section{Results}
\label{sec:result}
As shown in Table~\ref{table:longcontextResult}, we achieve state-of-the-art results on the PubMed and arXiv datasets. As shown in Table~\ref{table:newsResult}, our hierarchical seq2seq architecture outperforms its plain seq2seq peer initialized with the same pretrained weights on the news translation tasks XSum and CNN/DM and we achieve state-of-the-art results on the CNN/DM dataset. We outperform the previous baseline by 7 Rouge scores on SAMSum as shown in Table~\ref{table:conversationresult}. We also achieve state-of-the-art on ROUGE R2 on the AMI dataset, shown in Table~\ref{table:meetingresult}. File2rouge was used to evaluate Rouge score for all the results. Table~\ref{table:DocMT} shows our document level machine translation results. Our hierarchical seq2seq architecture outperforms the plain seq2seq baseline in English to German (En-De) translation, while for English to Czech (En-Cs) and Chinese to English (Zh-En) there is no clear improvement.
We hypothesize that this is because En-Cs and Zh-En contain significantly less documents than En-De (see Table~\ref{table:mt-datastats}) and the hierarchical layers are not able to make use of the BOS embeddings which are randomly
initialized.
\begin{table*}[h!]
\centering
\begin{tabular}{l c c c c c c} 
\toprule
& \multicolumn{3}{c}{\textbf{PubMed}} & \multicolumn{3}{c}{\textbf{arXiv}}\\
& R1 & R2 & RL & R1 & R2 & RL \\
\hline
PEGASUS~\cite{1912.08777} & 45.97 & 20.15 & 41.34 & 44.21 & 16.95 & 38.83 \\
BigBird~\cite{zaheer2020bigbird} & 46.32 & 20.65 & 42.33 & 46.63 & 19.02 & 41.77 \\ 
LSH~\cite{2104.02112} & 48.12 &21.06 &\textbf{42.72} & \textbf{48.24} & \textbf{20.26} & 41.78 \\
Transformer-BART & 48.35 & 21.43 & 36.90 & 46.54 & 18.82 & 42.00  \\ 
\hline
HAT-BART & \textbf{48.36} & \textbf{21.43} & 37.00 & 46.68 & 19.07 & \textbf{42.17} \\ 
\hline
\end{tabular}
\caption{Results on summarization tasks with long source sequences. PEGASUS~\cite{1912.08777} results are from BigBird~\cite{zaheer2020bigbird} paper. BigBird uses source sequence length of 4096, LSH~\cite{2104.02112} uses 7168, while Transformer-BART and HAT-BART use 3072  due to memory constraints. Transformer-BART and HAT-BART were trained using the same parameter settings.}
\label{table:longcontextResult}
\end{table*}

\begin{table*}[h!]
\centering
\begin{tabular}{l c c c c c c} 
\toprule
& \multicolumn{3}{c}{\textbf{CNN/DailyMail}} & \multicolumn{3}{c}{\textbf{XSum}}\\
& R1 & R2 & RL & R1 & R2 & RL \\
\hline
PEGASUS~\cite{1912.08777} & 44.16 & \textbf{21.56} & 41.30 & \textbf{47.60} & \textbf{24.83} & \textbf{39.64} \\ 
BigBird~\cite{zaheer2020bigbird} & 43.84 & 21.11 & 40.74 & 47.12 & 24.05 & 38.80 \\
BART~\cite{lewis2020bart} & 44.16 & 21.28 & 40.9 & 45.14 &	22.27	& 37.25 \\
Transformer-BART & 44.45	&21.27&	41.51 & 45.26&	22.19&	37.04 \\ 
\hline
HAT-BART & \textbf{44.48} & 21.31 & \textbf{41.52} & 45.92 & 22.79 & 37.84 \\ 
\hline
\end{tabular}
\caption{Results on standard news summarization tasks. PEGASUS results are from \citet{zaheer2020bigbird}. BigBird~\cite{zaheer2020bigbird}, Transformer-BART and HAT-BART use use a source sequence length of 1024. BART and HAT-BART were trained on the same parameters settings.}
\label{table:newsResult}
\end{table*}

\begin{table*}[h!]
\centering
\begin{tabular}{l c c c c c c c c c} 
\toprule
& \multicolumn{3}{c}{\textbf{SAMSum}}\\
& R1 & R2 & RL \\
\hline
DynamicConv + GPT2~\cite{gliwa-etal-2019-samsum} & 45.41 & 20.65 & 41.45 \\ 
Transformer-CNNDM & 53.00 & 28.03 & 48.59 \\ 
\hline
HAT-CNNDM & \textbf{53.01} & \textbf{28.27} & \textbf{48.84} \\ 
\hline
\end{tabular}
\caption{Results on conversational summarization tasks. The plain Transformer model (Transformer-CNNDM) was initialized by the BART CNN/DM model.
The hierarchical model (HAT-CNNDM) was initialized by the hierarchical seq2seq model trained on CNN/DM.}
\label{table:conversationresult}
\end{table*}

\begin{table*}[h!]
\centering
\begin{tabular}{l c c c c c c c c c} 
\toprule
& \multicolumn{3}{c}{\textbf{AMI}} & \multicolumn{3}{c}{\textbf{ISCI}}\\
& R1 & R2 & RL & R1 & R2 & RL \\
\hline
HMNet~\cite{zhu2020hierarchical} & \textbf{53.02} & 18.57 & - & \textbf{46.28} & 10.60 & - \\ 
Transformer-CNNDM & 52.06 & 19.27 & 50.02 & 43.77 & \textbf{11.65} & \textbf{41.64} \\ 
\hline
HAT-CNNDM & 52.27 & \textbf{20.15} & \textbf{50.57} & 43.98 & 10.83 & 41.36 \\ 
\hline
\end{tabular}
\caption{Results on meeting summarization tasks. The plain Transformer model (Transformer-CNNDM) was initialized by the BART CNN/DM model.
The hierarchical model (HAT-CNNDM) was initialized by the hierarchical Transformer model trained on CNN/DM.}
\label{table:meetingresult}
\end{table*}

\begin{table*}[h!]
\centering
\begin{tabular}{l c c c c c c} 
\toprule
& \textbf{WMT20} & \textbf{WMT19} & \textbf{WMT20} & \textbf{WMT19} & \textbf{TED17} & \textbf{TED16}\\
& \multicolumn{2}{c}{\textbf{En-De}} & \multicolumn{2}{c}{\textbf{En-Cs}} & \multicolumn{2}{c}{\textbf{Zh-En}}\\
\hline
Transformer (no pretrain) & 27.1 & 32.5 & \textbf{25.0} & 18.6 & 23.2 & \textbf{23.1}\\
\hline
HAT (no pretrain) & \textbf{27.4} & \textbf{34.5} & 24.2 & \textbf{19.2} & \textbf{24.0} & 22.9\\ 
\hline
\end{tabular}
\caption{Results on the WMT20/19 En-De, En-Cs and TED17/16 Zh-En translation tasks. The numbers shown are BLEU scores computed with sacrebleu.
The En-De and En-Cs models are trained on the WMT20 training data and we use the WMT18 test data for validation.
The Zh-En models are trained on the TED17 training data and we use the test data from TED11-TED15 and validation data from TED10
as validation data.
All models are initialized randomly with the same training setup.}
\label{table:DocMT}
\vspace{-1.4cm}
\end{table*}

\section{Analysis}
\label{sec:analysis}
The addition of hierarchical learning improves rouge scores over prior state-of-the-art methods for several summarization datasets.
Here we compare the generated output of our hierarchical model with the non-hierarchical counterpart for three instances from the arXiv test data.
We also include the introduction of each article. These can be found in Appendix~\ref{sec:appendix}.
Since there is often overlap between the abstract and the introduction, the models have learned to extract sentences from the introduction. We highlight the sentences extracted by the hierarchical model in blue.

\section{Ablation}\label{sec:ablation}

\paragraph{Encoder-only transformer hierarchical attention.}

\label{sec:encoderonly}
We evaluate our hierarchical attention model on several classification tasks. Instead of using our
seq2seq architecture, we design a similar encoder-only architecture with hierarchical attention. This also allows
us to easily pre-train the hierarchical layers.

Our architecture is based on the encoder of \cite{vaswani2017attention}. We add a hierarchical attention module after the self attention
module in each of the encoder layers. Similarly to how we preprocessed the summarization and translation datasets, we insert BOS tokens at the beginning
of sentences.

We follow RoBERTa~\cite{liu2019roberta} for pre-training our model by using the same dataset, pre-processing steps and pre-training objective.
We evaluate the pre-trained model on three downstream tasks:  SQuAD 2.0~\cite{rajpurkar2018know}, MNLI-m~\cite{WilliamsNB18} and RACE~\cite{LaiXLYH17}. 
We observe that the pre-training converges faster to a better optimum with lower complexity than RoBERTa with the same hyperparameters. However, downstream task performances are not improved consistently. 

The results are given in Table~\ref{table:encoder}. We observe that for SQuAD 2.0 and MNLI-m our hierarchical model does not perform better than the non-hierarchical model. However, the performance for RACE is significantly better, which suggests that there are some benefits in using hierarchical attention for classification tasks with long source sequences. Note that when fine-tuning on RACE, we had to disable dropout for the first epoch and then set it to 0.1, otherwise the model did not converge.

\vspace{-0.2cm}
\begin{table}[H]
\centering
\begin{tabular}{c c c c } 
\toprule
& {\small \textbf{SQuAD 2.0} } & {\small \textbf{MNLI-m}}  & {\small \textbf{RACE}} \\
& {\small F1} & {\small Acc} & {\small Acc}\\
{\small RoBERTa-Base$^1$} & 79.7 & 84.7 & 65.6 \\ 
{\small HAT (Encoder)} & 79.4 & 84.7 & 67.3 \\ 
\hline
\end{tabular}
\caption{Hierarchical learning for encoder only.}
{\tiny  $^1$ The results 
are taken from Table 1 (DOC-SENTENCES) in \citet{liu2019roberta}}
\label{table:encoder}
\end{table}
\paragraph{What has the hierarchical attention learned?}
 In order to better understand what the hierarchical model has learned, we plot a heatmap of the hierarchical attention between the decoder and the
 encoder. We use the best performing hierarchical model for this experiment (Table~\ref{table:longcontextResult}).
 We generate summaries for each sample article and record the hierarchical attention between the decoder and the BOS embeddings from the encoder at each step of generating the summary. For each of the 12 decoder layers and each of the 16 attention heads, we get a distribution over the BOS tokens for each generated summary tokens. To visualize the attention more concisely, we aggregate across the attention heads by choosing only the 16 BOS tokens with the highest weight for each generated token. We normalize
 the distribution such that the sum of the weights to 1.
The hierarchical attention heatmaps for each layer are shown in Figures 
\ref{figure:heatmaps-doc2} and \ref{figure:heatmaps-doc3}.
 
 We see that across different layers the model attends to different BOS tokens across the entire document. For each layer there are several
 horizontal lines, which indicates that some BOS tokens are assigned large weights at each step during generation. However, we also observe
 many individual points and discontinuous horizontal lines on the heatmaps, indicating that the model selectively attends to different
 BOS tokens across the entire document. We note that in the first few layers, the model seems to attend to the BOS tokens more uniformly
 while in the last few layers the model attends more heavily to certain BOS tokens.

\section{Conclusion}
\label{sec:conclusion}
We designed a transformer based architecture with hierarchical attention and obtained improvements on several sequence to sequence generation tasks. We showed significant improvements on document-level machine translation for the WMT20 En-De translation task, as compared to our baseline. We did not see significant gains when applying hierarchical attention
to encoder-only classification tasks.

\section{Future Work}
Although we did not see significant gains when using a hierarchical encoder-only
model for classification, we believe that a combination of modifying the architecture and the pre-training process  might improve over current non-hierarchical models for classification tasks with long source sequences. For instance, one might introduce
a pre-training objective specific to hierarchical attention such as predicting masked tokens in a sentence given only the
representation of the BOS token corresponding to that sentence.
We also believe the impact of dropout on hierarchical layers should be investigated more closely, since training on RACE diverged unless we disabled dropout during the first epoch (see Section~\ref{sec:ablation}).

\bibliographystyle{acl_natbib}
\bibliography{anthology,emnlp2020}

\newpage
\appendix
\onecolumn
\section{Appendix}
\label{sec:appendix}

\begin{table*}[hbt!]
\small
\centering
\begin{tabular}{p{\dimexpr \linewidth-2\tabcolsep}} 
\toprule
\multicolumn{1}{c}{arXiv test introduction - 2}\\
\hline\\
convolutional neural networks typically consist of an input layer , a number of hidden layers , followed by a softmax classification layer . the input layer , and each of the hidden layers , is represented by a three - dimensional array with size , say , @xmath0 . the second and third dimensions are spatial . the first dimension is simply a list of features available in each spatial location . for example , with rgb color images @xmath1 is the image size and @xmath2 is the number of color channels . the input array is processed using a mixture of convolution and pooling operations . as you move forward through the network , @xmath3 decreases while @xmath4 is increased to compensate . \hlblue{when the input array is spatially sparse , it makes sense to take advantage of the sparsity to speed up the computation . more importantly , knowing you can efficiently process sparse images gives you greater freedom when it comes to preparing the input data .}    consider the problem of online isolated character recognition ; \_ online \_ means that the character is captured as a path using a touchscreen or electronic stylus , rather than being stored as a picture . recognition of isolated characters can be used as a building block for reading cursive handwriting , and is a challenging problem in its own right for languages with large character sets . each handwritten character is represented as a sequence of strokes ; each stroke is stored as a list of @xmath5- and @xmath6-coordinates . we can draw the characters as @xmath1 binary images : zero for background , one for the pen color . the number of pixels is @xmath7 , while the typical number of non - zero pixels is only @xmath8 , so the first hidden layer can be calculated much more quickly by taking advantage of sparsity .    another advantage of sparsity is related to the issue of spatial padding for convolutional networks . convolutional networks conventionally apply their convolutional filters in \_ valid \_ mode  they are only applied where they fit completely inside the input layer . this is generally suboptimal as makes it much harder to detect interesting features on the boundary of the input image . there are a number of ways of dealing with this .    padding the input image @xcite with zero pixels . this has a second advantage : training data augmentation can be carried out in the form of adding translations , rotations , or elastic distortions to the input images .    adding small amounts of padding to each of the convolutional layers of the network ; depending on the amount of padding added this may be equivalent to applying the convolutions in \_ full \_ mode . this has a similar effect to adding lots of padding to the input image , but it allows less flexibility when it comes to augmenting the training data .    applying the convolutional network to a number of overlapping subsets of the image @xcite ; this is useful if the input images are not square . this can be done relatively computationally efficiently as there is redundancy in the calculation of the lower level convolutional filters . however , the ( often large ) fully connected classification layers of the network must be evaluated several times . sparsity has the potential to combine the best features of the above . the whole object can be evaluated in one go , with a substantial amount of padding added at no extra cost .    in section [ deepcnet]-[deepcnin ] \hlblue{we describe a family of convolutional networks with many layers of max - pooling .} in section  [ sparsity][nn ] we describe how sparsity applies to character recognition and image recognition . in section [ results ] we give our results . in section [ sec : conclusion ] we discuss other possible uses of sparse cnns .\\
\hline
\end{tabular}
\caption{Introduction of sample article from arXiv test data (2).}
\label{table:longcontextAnalysis-doc3}
\end{table*}

\begin{table*}[hbt!]
\small
\centering
\begin{tabular}{p{\dimexpr 0.1\linewidth-2\tabcolsep} p{\dimexpr 0.9\linewidth-2\tabcolsep}} 
\toprule
&\multicolumn{1}{c}{arXiv prediction and target - 2}\\
\hline\\
HAT & spatial sparsity is an important feature of convolutional neural networks ( cnns ) . when the input array is sparse , it makes sense to take advantage of the sparsity to speed up the computation . more importantly , knowing you can efficiently process sparse images gives you greater freedom when it comes to preparing the input data . we describe a family of cnn architectures with many layers of max - pooling , and show how sparsity can be used to improve the performance of online character recognition and image recognition .\\&\\
HAT2 & When the input array is spatially sparse, it makes sense to take advantage of the sparsity to speed up the computation.More importantly, knowing you can efficiently process sparse images gives you greater freedom when it comes to preparing the input data. consider the problem of online isolated character recognition; \_ online \_ means that the character is captured as a path using a touchscreen or electronic stylus, rather than being stored as a picture.Recognition of isolated characters can be used as a building block for reading cursive handwriting, and is a challenging problem in its own right for languages with large character sets.Each handwritten character is represented as a sequence of strokes; each stroke is stored as an array of coordinates.We show that using sparsity allows a substantial increase in the spatial resolution, allowing us to obtain good results for challenging datasets such as casia-olhwdb1. 1.\\&\\
Transformer & convolutional neural networks ( cnns ) are powerful tools for the recognition of spatially sparse objects . when the input is sparse , the network can be used to speed up learning , and the training data can be augmented to improve performance . we describe a family of cnn architectures with many layers of max - pooling , and use them to perform online and offline character recognition and image recognition .\\&\\
Target & convolutional neural networks ( cnns ) perform well on problems such as handwriting recognition and image classification . however , the performance of the networks is often limited by budget and time constraints , particularly when trying to train deep networks . motivated by the problem of online handwriting recognition , we developed a cnn for processing spatially - sparse inputs ; a character drawn with a one - pixel wide pen on a high resolution grid looks like a sparse matrix . taking advantage of the sparsity allowed us more efficiently to train and test large , deep cnns . on the casia - olhwdb1.1 dataset containing 3755 character classes we get a test error of 3.82\% . although pictures are not sparse , they can be thought of as sparse by adding padding . applying a deep convolutional network using sparsity has resulted in a substantial reduction in test error on the cifar small picture datasets : 6.28\% on cifar-10 and 24.30\% for cifar-100 . * keywords : * online character recognition , convolutional neural network , sparsity , computer vision\\
\hline
\end{tabular}
\caption{ROUGE(HAT): 37.25/12.25/34.01; ROUGE(Transformer): 36.84/7.96/31.58}
\label{table:longcontextAnalysis-doc4}
\end{table*}

\begin{table*}[hbt!]
\small
\centering
\begin{tabular}{p{\dimexpr \linewidth-2\tabcolsep}} 
\toprule
\multicolumn{1}{c}{arXiv test introduction - 3}\\
\hline\\
question answering ( qa ) aims to automatically understand natural language questions and to respond with actual answers . the state - of - the - art qa systems usually work relatively well for factoid , list and definition questions , but they might not necessarily work well for real world questions , where more comprehensive answers are required . frequently asked questions ( faq ) based qa is an economical and practical solution for general qa @xcite . instead of answering questions from scratch , faq - based qa tries to search the faq archives and check if a similar question was previously asked . if a similar question is found , the corresponding answer is returned to the user . the faq archives are usually created by experts , so the returned answers are usually of higher - quality . the core of faq - based qa is to calculate semantic similarities between questions . this is a very challenging task , because two questions , which share the same meaning , may be quite different at the word or syntactic level . for example ,  how do i add a vehicle to this policy ? " and  what should i do to extend this policy for my new car ? " have few words in common , but they share the same answer . in the past two decades , many efforts have been made to tackle this lexical gap problem . one type of methods tried to bridge the lexical gap by utilizing semantic lexicons , like wordnet @xcite . another method treated this task as a statistical machine translation problem , and employed a parallel question set to learn word - to - word or phrase - to - phrase translation probabilities @xcite . both of these methods have drawbacks . the first method is hard to adapt to many other languages , because the semantic lexicon is unavailable . for the second method , a large parallel question set is required to learn the translation probabilities , which is usually hard or expensive to acquire . to overcome these drawbacks , we utilize distributed word representations to calculate the similarity between words , which can be easily trained by only using amount of monolingual data . \hlblue{
in this paper , we propose a novel word - alignment - based method to solve the faq - based qa tasks . the characteristics of our method include : ( 1 ) a neural network model for calculating question similarity with word alignment features . for an input question and a candidate question , the similarities of each word pairs ( between the two questions ) are calculated first , and then the best word alignment for the two questions is computed . we extract a vector of dense features from the word alignment , then import the feature vector into a neural network and calculate the question similarity in the network s output layer . ( 2 ) a bootstrap - based feature extraction method . the faq archives usually contain less than a few hundred questions , and in order to avoid overfitting , we are unable to use too many sparse features . therefore , we come up with this method to extract a small set of effective sparse features according to our system s ranking results . ( 3 ) a learning - to - rank algorithm for training . the faq - based qa task is essentially a ranking task , our model not only needs to calculate a proper similarity for each question pair , but also needs to rank the most relevant one on top of the other candidates . so we propose a learning - to - rank method to train parameters more suitable for ranking . experimental results , conducted on faq archives from three languages , demonstrate that our method is very effective . we also evaluate our method on the answer sentence selection task . experimental results on the standard trec data set show that our method outperforms all previous state - of - the - art systems .}\\
\hline
\end{tabular}
\caption{Introduction of sample article from arXiv test data (3).}
\label{table:longcontextAnalysis-doc5}
\end{table*}

\begin{table*}[hbt!]
\small
\centering
\begin{tabular}{p{\dimexpr 0.1\linewidth-2\tabcolsep} p{\dimexpr 0.9\linewidth-2\tabcolsep}} 
\toprule
&\multicolumn{1}{c}{arXiv prediction and target - 3}\\
\hline\\
HAT & in this paper , we propose a novel word - alignment - based method to solve the frequently asked questions ( faq ) - based question answering ( qa ) tasks . the characteristics of our method include : ( 1 ) a neural network model for calculating question similarity with word alignment features . for an input question and a candidate question , the similarities of each word pairs ( between the two questions ) are calculated first , and then the best word alignment for the two question is computed . we extract a vector of dense features from the word alignment , then import the feature vector into a neural networks and calculate the question similarity in the network s output layer . ( 2 ) a bootstrap - based feature extraction method . the faq archives usually contain less than a few hundred questions , and in order to avoid overfitting , we are unable to use too many sparse features . therefore , we come up with this method to extract a small set of effective sparse features according to our system s ranking results . ( 3 ) a learning - to - rank algorithm for training . since faq - based qa task is essentially a ranking task , our model not only needs to calculate a proper similarity for each question pair , but also needs to rank the most relevant candidate on top of the other candidates . so we propose an explicit learning -to - rank method to train parameters more suitable for ranking . experimental results , conducted on faq archive from three languages , demonstrate that our method is very effective . we also evaluate our method on the answer sentence selection task , and experimental results on the standard trec data set show that our system outperforms all previous state - of - the - art systems .\\&\\
HAT2 &  In this paper, we propose a novel word-alignment-based method to solve the frequently asked questions (faq) based question answering (qa) tasks.The characteristics of our method include: (1) a neural network model for calculating question similarity with word alignment features. for an input question and a candidate question, the similarities of each word pairs (between the two questions)Are calculated first, and then the best word alignment for the two question is computed.We extract a vector of dense features from the word alignment, then import the feature vector into the neural network and calculate the question similarity in the network s output layer.(2) a bootstrap-based feature extraction method to extract a small set of effective lexical features according to the model s ranking results.(3) a learning-to-rank algorithm for training.The faq-based qa task is essentially a ranking task, our model not only needs to calculate a proper similarity for each question pair, but also needs to rank the most relevant candidate on top of the other candidates.So we propose to train parameters more suitable for ranking.Experimental results, conducted on faq archives from three languages, demonstrate that our method is very effective.We also evaluate our method on the answer sentence selection task.Experiment results on the standard trec data set show that our system outperforms all previous state-of-the-art systems. \\&\\
Transformer & in this paper , we propose a novel word - alignment - based method to solve the frequently asked questions ( faq ) based question answering ( qa ) tasks . the characteristics of our method include : ( 1 ) a neural network model for calculating question similarity with word alignment features . for an input question and a candidate question , the similarities of each word pairs ( between the two questions ) are calculated first , and then the best word alignment for the two question is computed . we extract a vector of dense features from the word alignment , then import the feature vector into a neural networks and calculate the question similarity in the network s output layer . ( 2 ) a bootstrap - based feature extraction method . the faq archives usually contain less than a few hundred questions , and in order to avoid overfitting , we are unable to use too many sparse features . therefore , we come up with this method to extract a small set of effective sparse features according to our system s ranking results . ( 3 ) a learning - to - rank algorithm for training . thefaq - based qa task is essentially a ranking task , our model not only needs to calculate a proper similarity for each question pair , but also needs to rank the most relevant one on top of the other candidates . so we propose an explicit learning - to - rank method to train parameters more suitable for ranking . experimental results , conducted on faq archive from three languages , demonstrate that our method is very effective . we also evaluate our method on the answer sentence selection task . experimental result on the standard trec data set shows that our system outperforms all previous state - of - the - art systems .\\&\\
Target & in this paper , we propose a novel word - alignment - based method to solve the faq - based question answering task . first , we employ a neural network model to calculate question similarity , where the word alignment between two questions is used for extracting features . second , we design a bootstrap - based feature extraction method to extract a small set of effective lexical features . third , we propose a learning - to - rank algorithm to train parameters more suitable for the ranking tasks . experimental results , conducted on three languages ( english , spanish and japanese ) , demonstrate that the question similarity model is more effective than baseline systems , the sparse features bring 5\% improvements on top-1 accuracy , and the learning - to - rank algorithm works significantly better than the traditional method . we further evaluate our method on the answer sentence selection task . our method outperforms all the previous systems on the standard trec data set .\\
\hline
\end{tabular}
\caption{ROUGE(HAT): 58.71/37.41/55.37; ROUGE(Transformer): 58.99/38.07/55.15}\label{table:longcontextAnalysis-doc6}
\end{table*}

\FloatBarrier
\section{Appendix}
\label{sec:appendix-c}

\begin{figure*}[h!]
\centering
\begin{subfigure}{\textwidth}
\centering
\includegraphics[width=\textwidth]{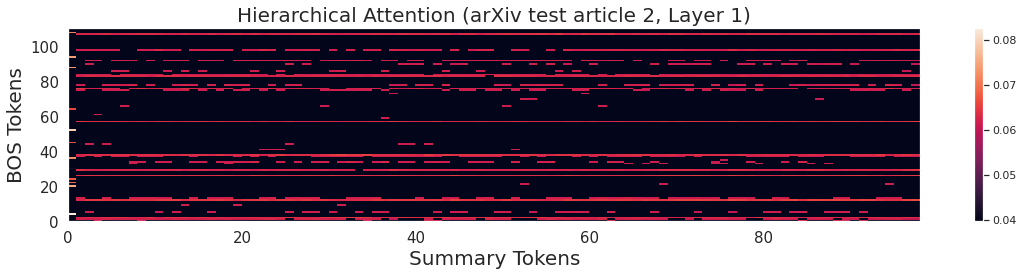}
\end{subfigure}
\begin{subfigure}{\textwidth}
\centering
\includegraphics[width=\textwidth]{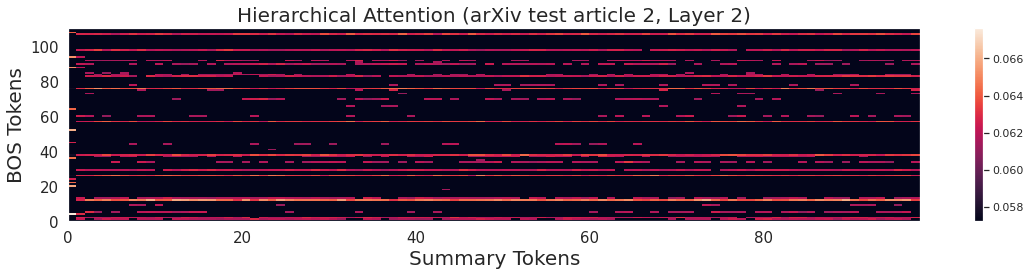}
\end{subfigure}
\begin{subfigure}{\textwidth}
\centering
\includegraphics[width=\textwidth]{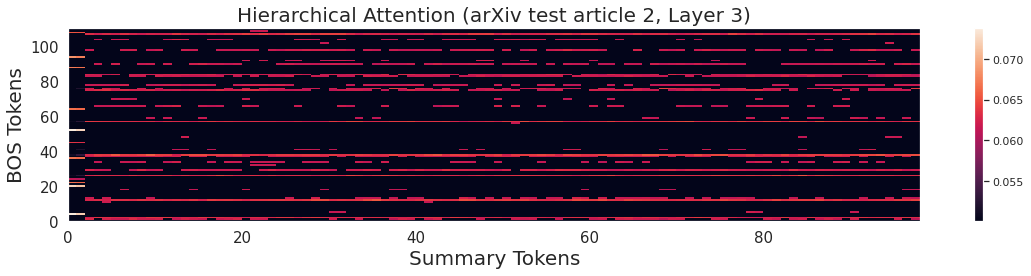}
\end{subfigure}
\end{figure*}%

\begin{figure*}[h!]\ContinuedFloat
\centering
\begin{subfigure}{\textwidth}
\centering
\includegraphics[width=\textwidth]{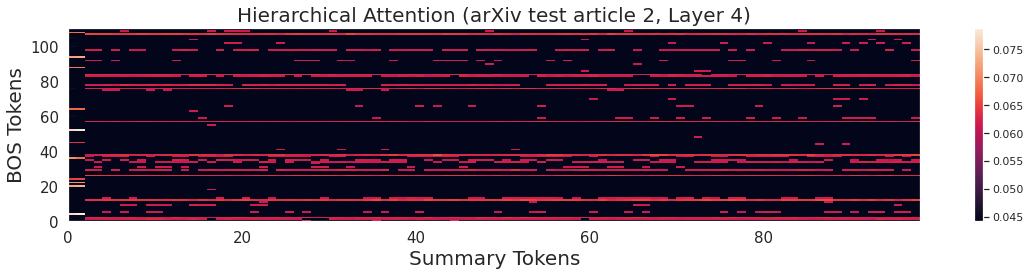}
\end{subfigure}
\begin{subfigure}{\textwidth}
\centering
\includegraphics[width=\textwidth]{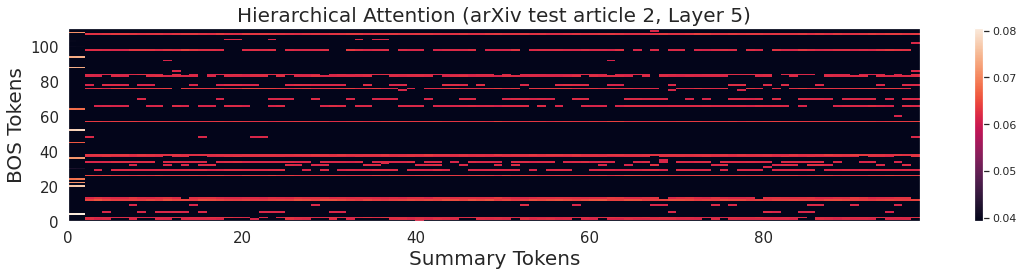}
\end{subfigure}
\begin{subfigure}{\textwidth}
\centering
\includegraphics[width=\textwidth]{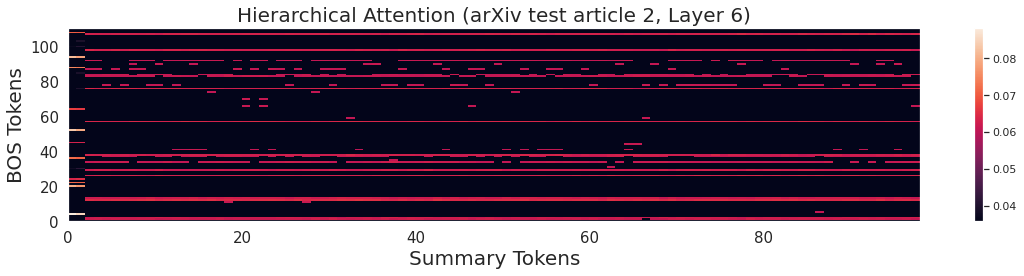}
\end{subfigure}
\begin{subfigure}{\textwidth}
\centering
\includegraphics[width=\textwidth]{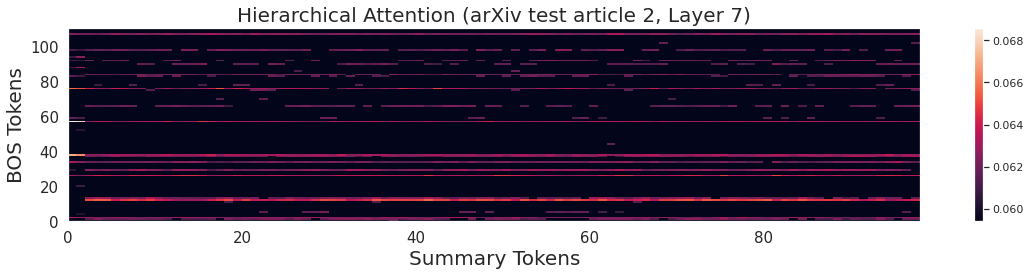}
\end{subfigure}
\begin{subfigure}{\textwidth}
\centering
\includegraphics[width=\textwidth]{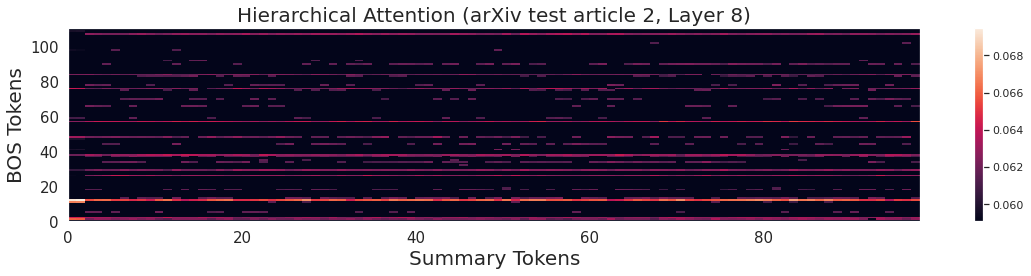}
\end{subfigure}
\end{figure*}%

\begin{figure*}[h!]\ContinuedFloat
\centering
\begin{subfigure}{\textwidth}
\centering
\includegraphics[width=\textwidth]{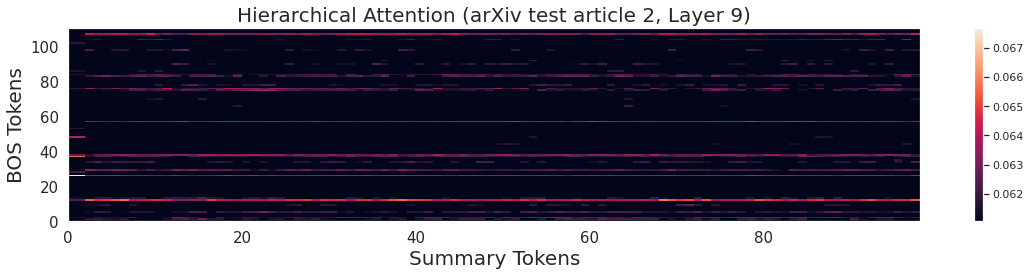}
\end{subfigure}
\begin{subfigure}{\textwidth}
\centering
\includegraphics[width=\textwidth]{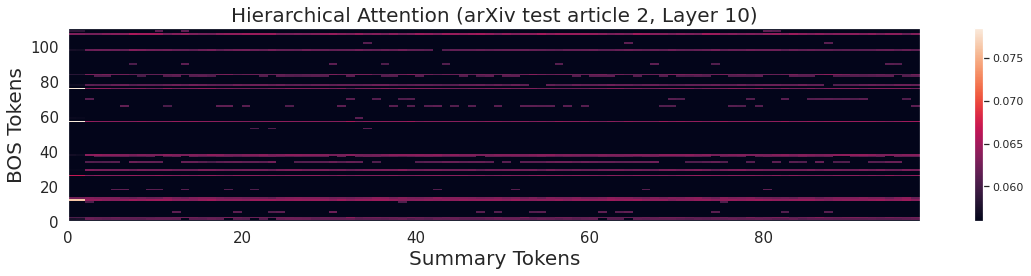}
\end{subfigure}
\begin{subfigure}{\textwidth}
\centering
\includegraphics[width=\textwidth]{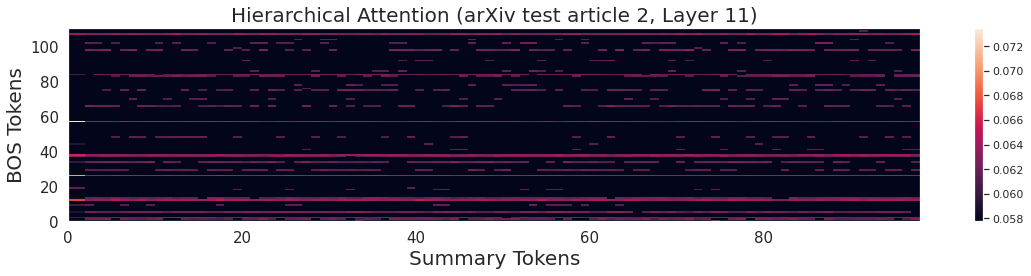}
\end{subfigure}
\begin{subfigure}{\textwidth}
\centering
\includegraphics[width=\textwidth]{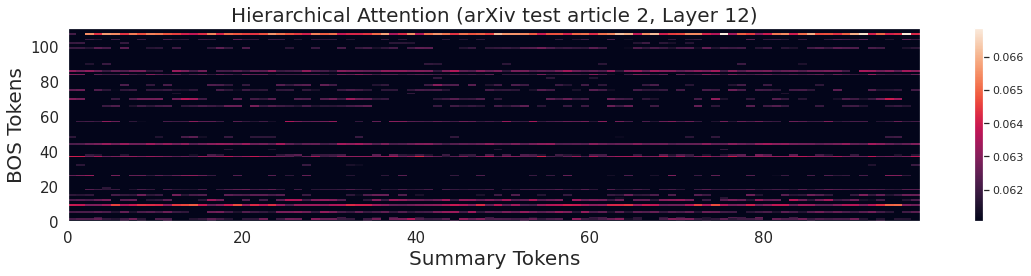}
\end{subfigure}
\caption{Hierarchical Attention heatmaps for arXiv test article 2 (Table \ref{table:longcontextAnalysis-doc3} and \ref{table:longcontextAnalysis-doc4}). For each summary token, we only show the top 16
BOS tokens across each head.}
\label{figure:heatmaps-doc2}
\end{figure*}


\begin{figure*}[h!]
\centering
\begin{subfigure}{\textwidth}
\centering
\includegraphics[width=\textwidth]{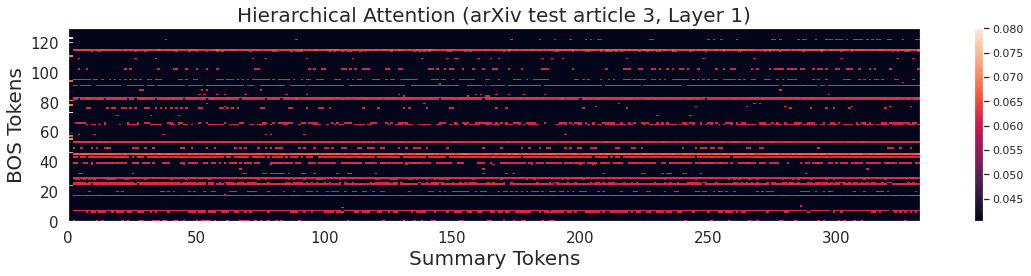}
\end{subfigure}
\begin{subfigure}{\textwidth}
\centering
\includegraphics[width=\textwidth]{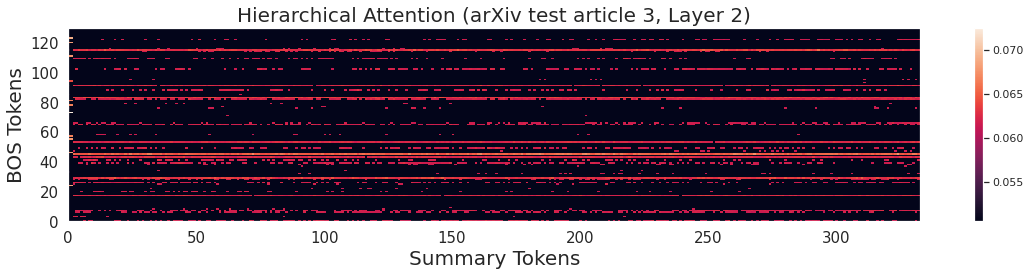}
\end{subfigure}
\begin{subfigure}{\textwidth}
\centering
\includegraphics[width=\textwidth]{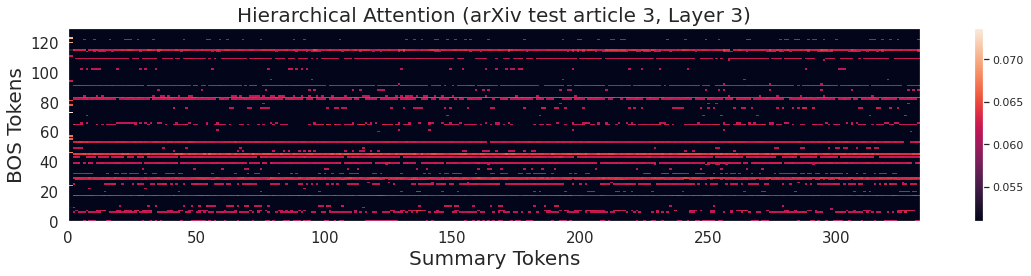}
\end{subfigure}
\begin{subfigure}{\textwidth}
\centering
\includegraphics[width=\textwidth]{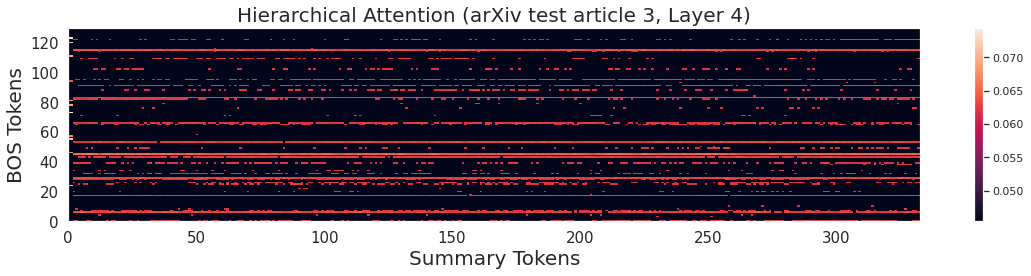}
\end{subfigure}
\begin{subfigure}{\textwidth}
\centering
\includegraphics[width=\textwidth]{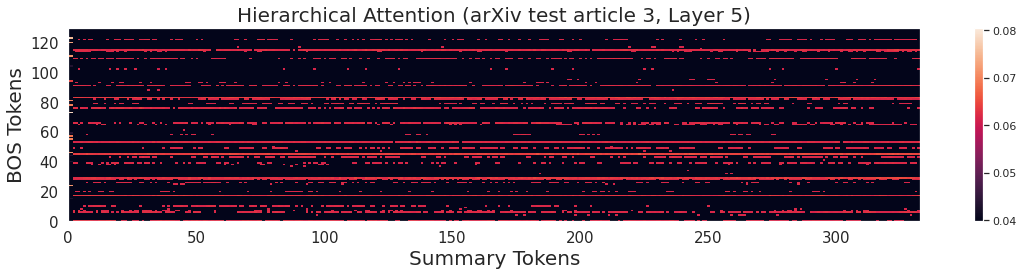}
\end{subfigure}
\end{figure*}%

\begin{figure*}[h!]\ContinuedFloat
\centering
\begin{subfigure}{\textwidth}
\centering
\includegraphics[width=\textwidth]{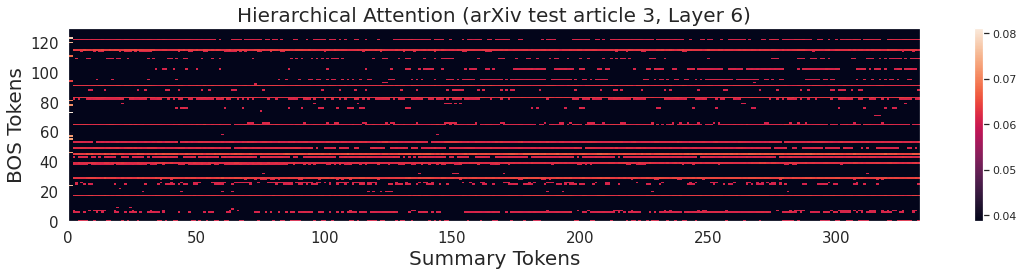}
\end{subfigure}
\begin{subfigure}{\textwidth}
\centering
\includegraphics[width=\textwidth]{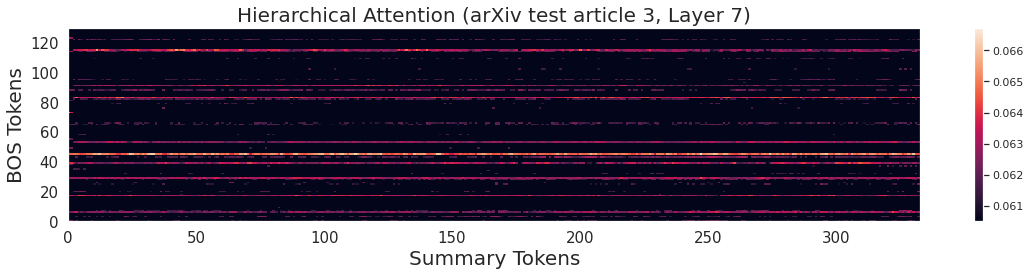}
\end{subfigure}
\begin{subfigure}{\textwidth}
\centering
\includegraphics[width=\textwidth]{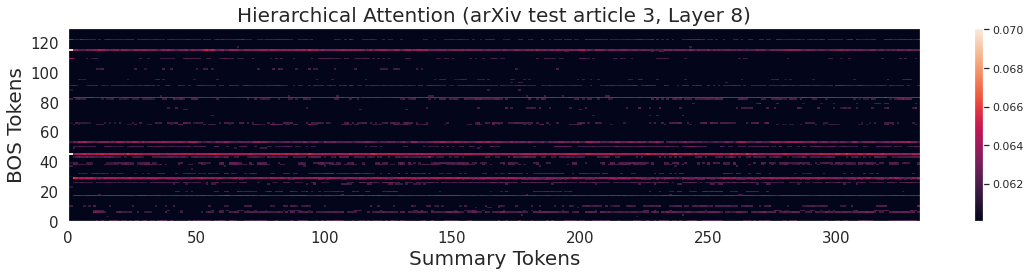}
\end{subfigure}
\begin{subfigure}{\textwidth}
\centering
\includegraphics[width=\textwidth]{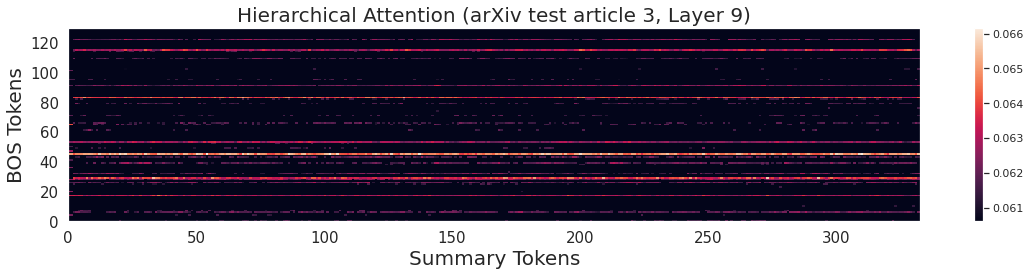}
\end{subfigure}
\begin{subfigure}{\textwidth}
\centering
\includegraphics[width=\textwidth]{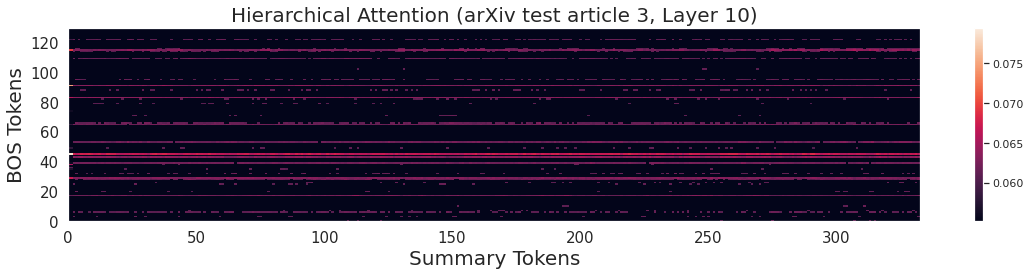}
\end{subfigure}
\end{figure*}%

\begin{figure*}[h!]\ContinuedFloat
\begin{subfigure}{\textwidth}
\centering
\includegraphics[width=\textwidth]{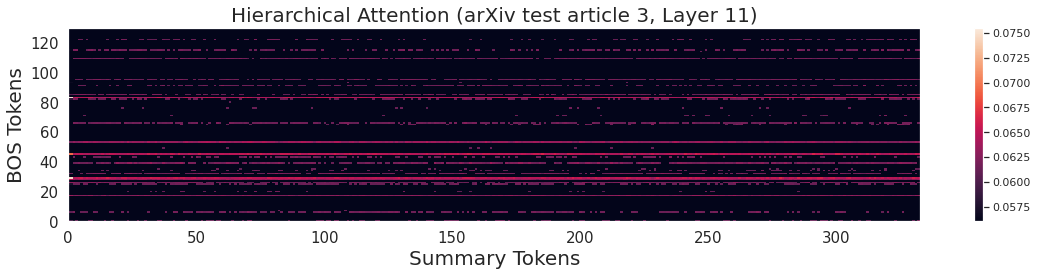}
\end{subfigure}
\begin{subfigure}{\textwidth}
\centering
\includegraphics[width=\textwidth]{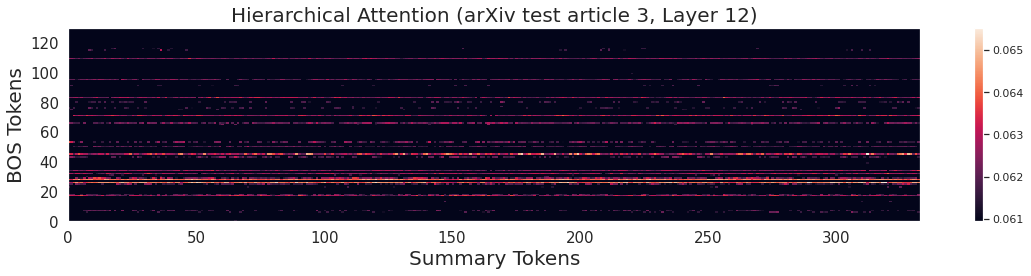}
\end{subfigure}
\caption{Hierarchical Attention heatmaps for arXiv test article 3 (Table \ref{table:longcontextAnalysis-doc5} and \ref{table:longcontextAnalysis-doc6}). For each summary token, we only show the top 16
BOS tokens across each head.}
\label{figure:heatmaps-doc3}
\end{figure*}

\end{document}